\documentclass[11pt]{article}
\pdfoutput=1
\usepackage[
  margin=1.5cm,
  includefoot,
  footskip=30pt,
]{geometry}
\counterwithin{figure}{section}
\usepackage{tikz}
\usetikzlibrary{matrix, arrows}
\usepackage{amsmath,amssymb}
\usepackage{amsthm}
\usepackage{mathtools}
\usepackage{xspace}
\usepackage[noend]{algorithmic}
\usepackage[ruled,vlined]{algorithm2e}
\usepackage{url}
\usepackage{makeidx}
\usepackage{enumerate}
\usepackage{epstopdf}
\usepackage{booktabs}
\usepackage{color}
\usepackage[utf8]{inputenc}
\usepackage{thm-restate}
\usepackage{scalerel,stackengine}
\usepackage[shortlabels]{enumitem}
\usepackage{xr}
\usepackage{fancyvrb}
\usepackage{xcolor}
\usepackage{bold-extra}
\usepackage[width=474.18663pt]{caption}
\usepackage{subfigure}
\usepackage[most]{tcolorbox}
\usepackage{fvextra}
\usepackage[frozencache=true, finalizecache=false, cachedir=./minted-cache]{minted} 
\usepackage{float}
\usepackage{alltt}
\usepackage{soul}
\usepackage{fancyvrb}
\usepackage{multirow}
\usepackage[final]{hyperref}
\usepackage[bottom]{footmisc}

\usepackage{cite} 
\usepackage{natbib}

\usepackage{listings}
\usepackage{tikz}
\usetikzlibrary{shapes,calc,positioning}

\global

\definecolor{myblue}{RGB}{30,113,223}

\tcbset{
  aibox/.style={
    width=474.18663pt,
    top=10pt,
    colback=white,
    colframe=black,
    colbacktitle=myblue,
    enhanced,
    center,
    attach boxed title to top left={yshift=-0.1in,xshift=0.15in},
    boxed title style={boxrule=0pt,colframe=white,},
  }
}

\tcbset{
  aibox1/.style={
    width=474.18663pt,
    top=10pt,
    colback=white,
    colframe=black,
    colbacktitle=black,
    center,
  }
}

\tcbset{
  aibox2/.style={
    width=474.18663pt,
    top=10pt,
    colback=white,
    colframe=black,
    colbacktitle=red,
    center,
  }
}

\newtcolorbox{AIbox}[2][]{aibox,title=#2,#1}
\newtcolorbox{AIbox1}[2][]{aibox1,title=#2,#1}
\newtcolorbox{AIbox2}[2][]{aibox2,title=#2,#1}

\definecolor{aigold}{RGB}{244,210, 1} 
\definecolor{aigreen}{RGB}{210,244,211} 

\sethlcolor{aigreen}

\definecolor{aired}{RGB}{255,180,181}

\newtcbox{\mybox}[1][green]{on line,
arc=0pt,outer arc=0pt,colback=#1!10!white,colframe=#1!50!black,
boxsep=0pt,left=0pt,right=0pt,top=0pt,bottom=0pt,
boxrule=0pt,bottomrule=0pt,toprule=0pt}

\newcommand{\mlogo}{\raisebox{-9pt}{\includegraphics[width=2.0em]{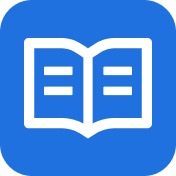}}\xspace\xspace}

\newcommand{\bx}{\boldsymbol{x}}
\newcommand{\bW}{\boldsymbol{W}}

\newcommand{\by}{\boldsymbol{y}}

\usepackage{authblk}

\makeatletter
\def\thanks#1{\protected@xdef\@thanks{\@thanks
        \protect\footnotetext{#1}}}
\makeatother

\begin{document}

\title{{\mlogo {AcademicGPT}: Empowering Academic Research}}

\author[$\dagger$]{Shufa Wei}
\author[$\dagger$]{Xiaolong Xu}
\author[$\dagger$]{Xianbiao Qi\thanks{$\dagger$ denotes equal contributions.}}
\author[$\dagger$]{Xi Yin}
\author[ ]{Jun Xia}
\author[ ]{Jingyi Ren}
\author[ ]{Peijun Tang}
\author[$\ast$]{\\Yuxiang Zhong\thanks{$\ast$~Work was done when they worked at IDEA.}}
\author[ ]{Yihao Chen}

\author[ ]{Xiaoqin Ren}
\author[ ]{Yuxin Liang}
\author[$\ast$]{Liankai Huang}

\author[ ]{Kai Xie}
\author[ ]{Weikang Gui}
\author[ ]{\\Wei Tan}

\author[ ]{Shuanglong Sun}
\author[ ]{Yongquan Hu}
\author[ ]{Qinxian Liu}
\author[ ]{Nanjin Li}

\author[ ]{Chihao Dai}

\author[ ]{Lihua Wang}

\author[$\ddagger$]{\\Xiaohui Liu}
\author[$\ddagger$]{Lei Zhang}
\author[$\ddagger$]{Yutao Xie\thanks{$\ddagger$  denotes equal leadership.}}

\affil[ ]{\Large \textcolor{blue}{ReadPaper Team}}
\affil[ ]{\LARGE \textcolor{blue}{International Digital Economy Academy}}

\date{\vspace{-5ex}}

\maketitle

\begin{abstract}
Large Language Models (LLMs) have demonstrated exceptional capabilities across various natural language processing tasks. Yet, many of these advanced LLMs are tailored for broad, general-purpose applications. In this technical report, we introduce AcademicGPT, designed specifically to empower academic research. AcademicGPT is a continual training model derived from LLaMA2-70B. 
Our training corpus mainly consists of academic papers, thesis, content from some academic domain, high-quality Chinese data and others. While it may not be extensive in data scale, AcademicGPT marks our initial venture into a domain-specific GPT tailored for research area.
We evaluate AcademicGPT on several established public benchmarks such as MMLU and CEval, as well as on some specialized academic benchmarks like PubMedQA, SCIEval, and our newly-created ComputerScienceQA, to demonstrate its ability from general knowledge ability, to Chinese ability, and to  academic ability. 
Building upon AcademicGPT's foundation model, we also developed several applications catered to the academic area, including General Academic Question Answering, AI-assisted Paper Reading, Paper Review, and AI-assisted Title and Abstract Generation.
\end{abstract}



\section{Introduction}
\label{sec:introduction}
Large Language Models (LLMs)~\citep{gpt_radford2018improving, gpt2_radford2019language, gpt3_brown2020language, instructgpt_ouyang2022training, openai_gpt4, palm_chowdhery2022palm, chinchilla_hoffmann2022training, chatglm_du2021glm, chatglm_130b_zeng2022glm, llama_touvron2023llama, llama2_touvron2023llama} have transformed natural language processing (NLP) and artificial intelligence (AI). LLMs have not only redefined our capabilities in understanding and generating text content but have also branched their influence into various domains. In the sphere of writing, LLMs~\citep{openai_gpt4} have breathed life into nuanced narratives and precise technical content. In the programming world, they~\citep{codellama_roziere2023code} have offered solutions to intricate coding problems, bridging the gap between human language and code. Moving to sectors like finance, these models~\citep{bloomberggpt_wu2023bloomberggpt} decode complex datasets and predict market trends with precision. In the healthcare domain~\citep{medpalm_singhal2022large, openai_gpt4, openai2023gpt4vision}, they assist in diagnosis, treatment suggestions, and even complex research tasks. In the creative arts, combining with multi-modality large models, they have opened doors to AI-driven music generation, costume designing
, and other forms of artistic expression. 
In conclusion, LLMs have revolutionized numerous industries~\citep{codegeex_zheng2023codegeex, codellama_roziere2023code, chemcrow_bran2023chemcrow, gpt_pinn_chen2024gpt, mathgpt_scarlatos2023tree, medpalm_singhal2022large, bloomberggpt_wu2023bloomberggpt, chatlaw_cui2023chatlaw, biogpt_luo2022biogpt}.

However, the journey to this revolutionary phase has not happened overnight. BERT~\citep{bert_devlin2018bert}  and GPT-1~\citep{gpt_radford2018improving} ushered in the era of large models. Models like GPT-3~\citep{gpt3_brown2020language} laid the groundwork, with its billions of parameters setting new benchmarks. Subsequent innovations, including ChatGPT's~\citep{instructgpt_ouyang2022training} conversational prowess, PalM's~\citep{palm_chowdhery2022palm} multitasking abilities, the LLaMA series'~\citep{llama_touvron2023llama, llama2_touvron2023llama} advanced linguistic capabilities, CodeGeeX and CodeLLaMA's~\citep{codegeex_zheng2023codegeex, codellama_roziere2023code} programming ability,
GPT4's~\citep{openai_gpt4} improved general and professional ability, and GPT-4V~\citep{openai2023gpt4vision, gpt4dawn_yang2023dawn}'s multi-model ability have continuously pushed the envelope, setting new frontiers in what AI can achieve. In conclusion, thanks to the ease of information dissemination, the pace of innovation has significantly outstripped that of the past. 


With knowledge burgeoning and scientifical discoveries emerging at an astonishing rate, scholars and researchers are continually overwhelmed by an expanding ocean of literature.  This overwhelming abundance is paradoxical, signifying both our triumphant strides in human understanding and the looming challenge that researchers face in keeping abreast of fresh insights. This issue becomes especially pronounced within specialized sectors or subdivisions. Here, the rapid growth of targeted studies, novel methodologies, and intricate findings intensifies the difficulty for scholars to rapidly understand and assimilate the particulars of these niche domains. Such information saturation hampers not only the smooth flow of knowledge but also erects barriers for interdisciplinary endeavors. Grasping the intricate details of these subdivisions demands significant time, slowing down the pace of integration and innovation.

\begin{figure}[t]
\centering
\includegraphics[width=1.0\textwidth]{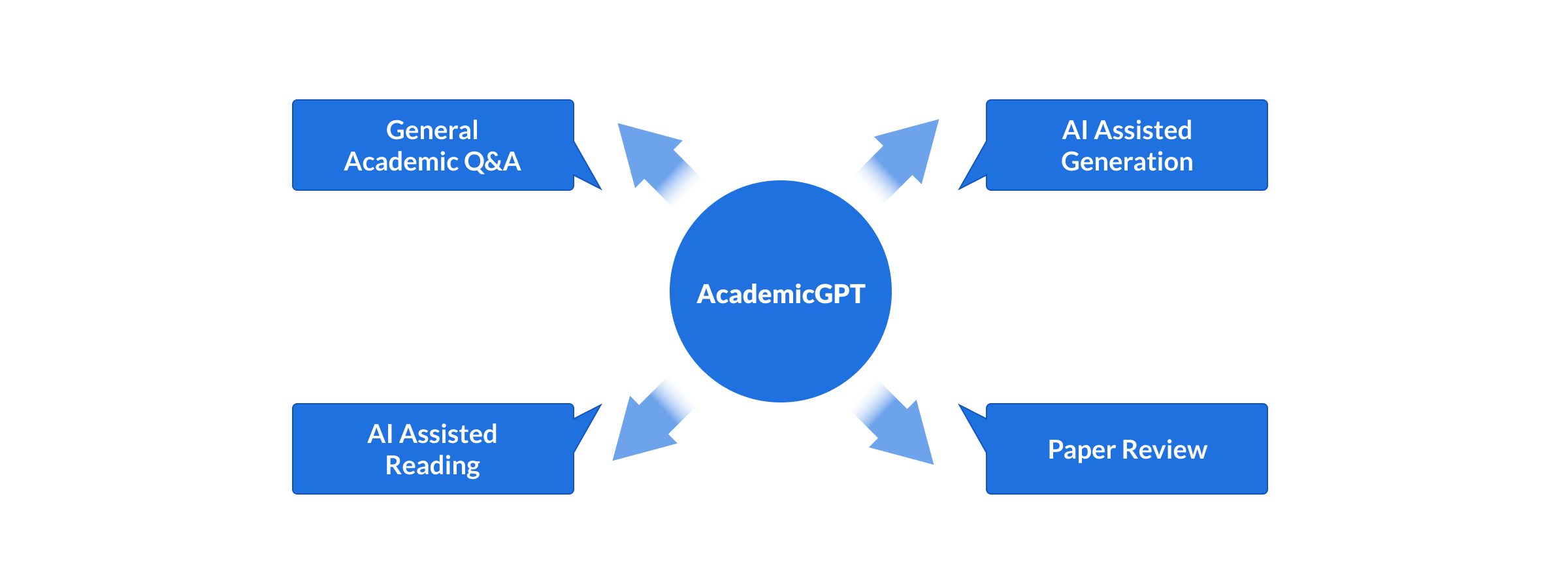}
\caption{An overview of AcademicGPT and four applications around it.}
\label{fig:overall_academicgpt_and_applications}
\end{figure}

It becomes increasingly crucial to provide researchers with effective tools and methodologies that allow them to distill essential insights from the vast ocean of information, ensuring that critical advancements and findings are recognized and built upon. These tools are not just limited to aiding in comprehension but span a broad spectrum of research activities, including \textit{paper reading} where AI-assisted methods can highlight key findings and offer a concise summary, \textit{paper polishing} where advanced tools can provide grammar checks, stylistic recommendations, and ensure the clarity and coherence of the presented ideas, \textit{paper reviewing} where tool can give a critical comments about the paper, 
\textit{content-based paper writing} where predictive and generative models can assist researchers in constructing well-structured narratives and arguments, saving them invaluable time.

In this technical report, our contributions are highlighted as two folders. 
\begin{itemize}[leftmargin=*]
    \item  we introduce AcademicGPT, a GPT model specifically tailored for scientific research. This model stands as a testament to the power of harnessing vast academic corpora, having been trained on a academic corpus with 120 billion tokens. The sheer volume of data processed ensures its robustness and accuracy in comprehending intricate scientific nuances. 
   
    \item we build several applications based on AcademicGPT, as shown in Figure~\ref{fig:overall_academicgpt_and_applications}, including General Academic Question Answering, AI-Assisted Paper Reading, Paper Review and AI-assisted Content Generation.
   Our General Academic Q\&A system is a sophisticated agent equipped with multi-turn dialogue memory. In the agent, our strategic planning and application architecture draw inspiration from the ReAct framework, integrating its principles to achieve the desired outcomes.
   This ensures continuity and context-awareness in academic discussions, setting the stage for meaningful and deep interactions. Recognizing the challenges presented by lengthy academic articles, we introduced an AI-powered solution to simplify and enhance the paper reading experience, ensuring researchers grasp the core concepts efficiently.
   Our paper review system, underpinned by the supervised finetuning (SFT) model based on AcademicGPT, introduces a way of assessing academic content. Our AI-Powered Content Generation  generate content such as abstracts and titles based solely on a given introduction. By manipulating the order of input context, our model exhibits strong adaptability in content creation.
\end{itemize}
In essence, our work with AcademicGPT not only introduces a powerful model for scientific research but also demonstrates its practical applications, promising a transformative impact on the academic community.

The structure of this technical report is depicted as follows: Section~\ref{sec:related_work} discuss some related works. Section~\ref{sec:methods} describes the AcademicGPT model and report its results on several benchmarks. Section~\ref{sec:applications} describes four applications built on AcademicGPT.

\

\section{Related Work}
\label{sec:related_work}

{\bf{Large Language Models (LLMs)}.} 
The domain of natural language processing (NLP)~\citep{bert_devlin2018bert, gpt_radford2018improving} and artificial intelligence (AI)~\citep{openai2023gpt4vision, gpt4dawn_yang2023dawn} has witnessed a transformative shift, primarily driven by the emergence and rapid evolution of LLMs~\citep{elmo_peters2017semi, bert_devlin2018bert, neox_black2022gpt, gpt_radford2018improving, gpt2_radford2019language, gpt3_brown2020language}. These models, with their unprecedented scale and capability, have redefined the paradigms of linguistic understanding, reasoning, and generation.
From a historical perspective, the journey of LLMs began with models comprising millions of parameters, like  GPT-1~\citep{gpt_radford2018improving}. However, as the field matured, the scale expanded drastically, moving to models boasting billions, or even trillions, of parameters, such as GPT-3~\citep{gpt3_brown2020language}, PaLM~\citep{palm_chowdhery2022palm}, and Switch Transformers~\citep{switch_transformer_fedus2022switch}. This massive increase in model size has been a cornerstone in enhancing their capabilities, offering more human-like fluency and versatility in a plethora of natural language tasks.
Interestingly, two main trajectories have dominated the LLM landscape: \textit{closed-source models} and \textit{open-source models}.
\textit{Closed-source Models}  such as GPT-4~\citep{openai_gpt4, gpt4dawn_yang2023dawn, agi_bubeck2023sparks}, ChatGPT~\citep{instructgpt_ouyang2022training}, Claude~\citep{anthropic_bai2022training}, PaLM~\citep{palm_chowdhery2022palm}, Chinchilla~\citep{chinchilla_hoffmann2022training}, Gopher~\citep{gopher_rae2021scaling} and ERNIE~\citep{ernie_zhang2019ernie, ernie30_sun2021ernie} have a dominating position in current LLM research and applications. Their introduction has reshaped the general perception about machine capabilities. For instance, ChatGPT's capacity to engage in diverse linguistic interactions, ranging from casual dialogues to elucidating intricate topics, underscores the potential of LLMs in automating tasks requiring linguistic prowess. However, a significant drawback accompanying these closed-source behemoths like GPT-4, PaLM-2, and Claude is the restricted access to their full parameters. This limitation hampers the broader research community from delving deep into these systems or optimizing them further, thereby constraining a collective progress.
In contrast to their closed-source counterparts, \textit{open-source models} such as OPT~\citep{opt_zhang2022opt}, Bloom~\citep{bloom_scao2022bloom}, Falcon~\citep{falcon_almazrouei2023falcon}, Baichuan~\citep{baichuan2_yang2023baichuan},  QWen~\citep{qwen_technical_report}, LLaMA1~\citep{llama_touvron2023llama} and LLaMA2~\citep{llama2_touvron2023llama} champion the cause of transparency and community engagement. LLaMA1, for example, with its vast 65 billion parameters, is not just a marvel in itself but also an exemplar of openness. The full availability of such models has been a boon, as researchers and developers can probe, experiment, and build upon them without constraints. This liberal approach has acted as a catalyst, furthering research and leading to the birth of new models like Alpaca~\citep{alpaca}, Vicuna~\citep{vicuna2023}, and more.
As the field forges ahead, it remains to be seen how these two pathways coalesce or diverge, but what is undeniable is their collective contribution to the magnificent world of LLMs.

\noindent
{\bf{Continual Pretraining of LLMs}.} Continual pretraining refers to the process of incrementally and continuously training a model on new data. Roughly speaking, 
continual pretraining can be categorized into four classes that targets different goals including \textit{lifelong pretraining, 
longer context window, domain adaptive learning, improving training strategy of continual pretraining.}
In the realm of \textit{lifelong learning}, \citet{lifelong_jin2021lifelong} introduced the concept of ``lifelong pretraining''. They explore where various continual learning algorithms were employed to incrementally pretrain language models. Through evaluations on the model's adaptability to new data, they find that distillation-based approaches effectively preserve the performance on downstream tasks from earlier domains. 
In another noteworthy work, \citet{elle_qin2022elle} proposed a model named ELLE. This model aspires to achieve efficient lifelong pretraining by leveraging pre-trained language model (PLM) extensions and pretraining domain prompts. Its primary aim is to adapt to continuously streaming data. 
In the domain of \textit{expanding context window}, several key works have made noteworthy contributions. \citet{codellama_roziere2023code} employed continual pretraining to enlarge the model's window. \citet{long_xiong2023effective} further advanced this paradigm, achieving a series of long-context LLMs that support effective context windows of up to 32,768 tokens, starting from a foundation of continual pretraining on LLaMA2. Targeting the RoPE~\citep{rope_su2021roformer} positional encoding, \citet{yarn_peng2023yarn} introduced the YaRN approach, a novel methodology devised specifically for expanding the model's context window.
In domain of \textit{adaptive learning}, there have been several pivotal contributions. \citet{codellama_roziere2023code} leveraged a continual pretraining to enhance the code capabilities of LLMs. They introduced, ``CodeLlama'', a state-of-the-art large-scale code language model built upon LLaMA2, boasting unrivaled performance in the open-source community, exceptional code completion capabilities, support for extensive input context sizes, and the adeptness in autonomously following directives in programming tasks. Furthermore, \citet{domain_zhang2023large} delved into the domain of continual pretraining in the context of biomedical visual language processing, shedding light on the nuances of domain-specific adaptations.
In the domain of \textit{improving training strategy} with a focus on stability, a series of studies have been conducted to advance our understanding. \citet{warm_gupta2023continual} explored various training approaches for continual pretraining. They examined the effects of different warming-up strategies on large language models. Their findings highlighted that restarting model warm-up can boost downstream performance, even outperforming models trained from scratch on sizable downstream datasets. 
In a parallel vein, \citet{continual_ke2022continual} investigated methodologies to enhance performance in domain-specific scenarios via continual pretraining. Their research proposed an innovative technique that harnesses a series of unlabeled domain-specific corpora for the continual pretraining of language models, thereby augmenting their end-task efficacy.

\noindent
{\bf{Domain-Specific  LLMs}.} LLMs have been applied to different domains after its success in natural language processing and AI. 
In \textit{scientific research}, Galactica~\citep{galactica_taylor2022galactica} model stands out as a tool tailored for general scientific research, streamlining the process of inquiry and discovery in the vast expanse of scientific literature. In the domain of \textit{Finance}, BloombergGPT~\citep{bloomberggpt_wu2023bloomberggpt} is tailored for the financial sector, providing insights, analyses, and information tailored to financial professionals and stakeholders. In \textit{Medicine}, Med-PaLM~\citep{medpalm_singhal2022large} is engineered specifically for the medical domain, ensuring accurate and context-aware responses pertinent to medical professionals and researchers. In \textit{Programming}, Code-LLaMA~\citep{codellama_roziere2023code} is geared towards programming, aiding developers by understanding and generating code, making the coding process more intuitive and efficient.
In \textit{Legal}, ChatLaw~\citep{chatlaw_cui2023chatlaw} emerges as an open-source legal LLM, providing a new way legal professionals access, interpret, and utilize legal texts.
In Biomedical, with models like BioGPT~\citep{biogpt_luo2022biogpt}, the biomedical domain can benefit from advanced text generation and mining, aiding in research, diagnosis, and treatment planning.
In \textit{Physics}, GPT-PINN~\citep{gpt_pinn_chen2024gpt} is a confluence of physics and AI, designed as a Physics-Informed Neural Network. It is tailored for meta-learning of parametric PDEs, offering a non-intrusive approach to solving complex physics problems.
In \textit{Mathematics}, MathGPT~\citep{mathgpt_scarlatos2023tree} targets the realm of mathematical reasoning, assisting researchers and students in understanding complex mathematical concepts and problems.

In summary, the rise of domain-specific LLMs underscores the potential of AI to cater to specialized needs across diverse fields. These models not only amplify the capabilities within their respective domains but also promise to transform the way professionals across sectors approach and solve challenges.

\ 

\noindent
\textbf{Remark}. AcademicGPT builds upon the foundation of LLaMA2, an open-source Large Language Model (LLM) renowned for its versatility and extensive capabilities. AcademicGPT is a continual pretraining on LLaMA2.
The primary domain of focus for AcademicGPT is academic research, our initial motivation is inspired by Galactica~\citep{galactica_taylor2022galactica}. AcademicGPT marks our initial venture into a domain-specific GPT tailored for research area.
In essence, AcademicGPT targets to help researchers, academicians, and students to quickly understand the fresh insights.

\ 

\section{AcademicGPT}
\label{sec:methods}
In this section, we delve into AcademicGPT by examining its data sources, model architecture, and experimental results. We begin by elucidating the datasets that are employed to cultivate AcademicGPT's capabilities. Then we give an overview of the model's architecture. We conclude by reported the model's performance on benchmarks such as MMLU~\citep{mmlu_hendrycks2020measuring}, CEval~\citep{ceval_huang2023c}, PubMedQA~\citep{pubmedqa_jin2019pubmedqa}, SCIEval~\citep{scieval_sun2023scieval}, and our newly collected ComputerScienceQA.

\ 

\subsection{Data}
Our goal in AcademicGPT is to enhance LLaMA2's capcibility on academic research, and meanwhile improve its ability on Chinese language. Therefore, the way of our data collection is around this two targets. It is well known that the capcibility of LLaMA2 on understanding Chinese language is limited due to its limited usage of Chinese corpus. Meanwhile, since LLaMA2 is a general LLM, that does not use enough academic data. Some existing large-scale dataset includes the Pile~\citep{pile_gao2020pile}, Roots~\citep{roots_laurenccon2022bigscience}, RedPajama-Data~\citep{redpajama_together2023redpajama}, Falcon-Refinedweb~\citep{refinedweb_penedo2023refinedweb}, WudaoCorpora Text~\citep{wudao_yuan2021wudaocorpora}. These data were collected for general purpose.

Our training data is constructed based on the up-mentioned two goals. 
\begin{itemize}
    \item including more academic data.
    \item adding more Chinese data
\end{itemize}
Specifically, on one side, our training data should consists of both high-quality Chinese and english data. On another side, our training data should be mainly from academic area, including academic paper, thesis, content from some academic domains, and more.

Our Chinese data consists of four types: Common Crawl (CC), Wiki, Baike, and Books. However, the data collected from CC are usually very dirty, it includes a lot of advertisements, pornographic information, violence
and other toxic information. We need to clean the data. 
Our Chinese data cleaning pipeline includes four stages. 1) we crawl 200K of articles from some top academic domains; (2) we use a powerful LLM to label the data.  The prompt we use to label the data is shown as in Figure~\ref{fig:label_data_prompt}; (3) we train a LLM for classification using the obtained 200K data; (4) we conduct classification on the common crawl data and clean out 12.7B tokens from CC.

\begin{table}[H]
    \centering
    \begin{tabular}{c|c|c|c}
         \hline
         Data Type & Data Source & Number of Tokens (Billions) & Sampled Rate\\ \hline
\text { Thesis } & Top Universities & 9.31 B & 1.0 \\ 
\text { Papers } & Arxiv & 85.52 B  & 0.5 \\ 
\text { Papers   } & Unpaywall & 208.22 B & 0.15 \\ 
\text { Papers } & PubMed & 20.79 B  & 0.5 \\ 
\text { Bibliographic } & Semantic Scholar & 22.14 B  & 0.3 \\ 
\text { Academic Domains } & Falcon RefinedWeb & 7.15 B  & 0.8 \\ 
\text { Chinese data } & Common Crawl & 12.76 B  & 1.0\\ 
\text { Wiki (English) } & Wiki & 4.96 B  & 0.5 \\  
\hline 
All & -  & 370.85 B  & 122.3 B \\
\hline
    \end{tabular}
    \caption{Data for our continual pre-training.}
    \label{tab:training_data}
\end{table}


\begin{figure}[H]
\begin{AIbox1}{{Prompt to label data \xspace}}

\parbox[t]{1.0\linewidth}{
\tiny
\begin{alltt} 
{

SYS\_PROMPT = ``In terms of checking data quality, you are a helpful and precise assistant.'' \\
PROMPT = """ \\
Please assess the provided CommonCrawl sample based on the following criteria and return the results in the specified JSON format.\\
 
The evaluation criteria are as follows:\\
1. Quality: Evaluate grammar completeness, language coherence, information accuracy, and the presence of low-quality content such as explicit, violent, advertising, promotional, or recruitment-related information. Categorize quality as ``Excellent'', ``Average'', or ``Poor'';\\

2. Domain: Determine if the sample is related to fields such as computer science, natural sciences, social sciences, engineering and technology, medical and health, arts and literature, humanities, economics and management, law, education, agricultural sciences, space sciences, etc., and categorize it accordingly;\\

3. Depth: Assess the content as ``Beginner'', ``Intermediate'', ``Advanced'', or ``Expert'';\\

4. Category: Identify whether it falls under the category of ``Academic Article'', ``Academic Report'', ``Monograph'', ``Whitepaper'', ``Technical Blog'', ``Popular Science Article'',``Forum Discussion'',``News Report'', or ``Promotional Content'';\\

5. Suitability Rating: Determine whether the sample is suitable for training academic large models, imparting serious knowledge to the model, enhancing the model's academic, common-sense, logical, and reasoning capabilities. Suitability has three standards: ``Highly Suitable'', ``Average'', or ``Not Suitable''.\\
 
The returned results should be in the following format:\\
\{\\
Quality: Excellent/Average/Poor, \\
Domain: Computer Science/Natural Sciences/Social Sciences/Engineering and Technology/Medical and Health/Arts and Literature/Other/Promotional Content, \\
Depth: Beginner/Intermediate/Advanced/Expert, \\
Category: Academic Article/Academic Report/Monograph/Whitepaper/Technical Blog/Popular Science Article/Forum Discussion/News Report/Promotional Content/Other, \\
Suitability: Highly Suitable/Average/Not Suitable\\
\}.\\

Please note that you only need to directly return the JSON results without providing any additional unnecessary text.""" \\


}
\end{alltt}
}
\end{AIbox1}
\caption{Prompt to label data.}
\label{fig:label_data_prompt}
\end{figure}

When collecting Academic English data, we focus on collecting a more high-quality  data. Our academic data consists of a mix of several sources. First, we crawl more than 1 million of theses from 200 top universities in the world. We believe the thesis data, compared to traditional conference or journal paper, are self-consistent. Instead, conference and journal papers are usually inconsistent, and requires more expert experience to understand. Since the content in the thesis is usually very long. For this kind of data, we use Nougat~\citep{nougat_blecher2023nougat} to parse the pdf files. Second,
we crawl Arxiv~\footnote{\url{https://arxiv.org/}} papers (it contains around 2.26 millions of papers until to May 2023.). Third, we use the data from unpaywall~\footnote{\url{https://unpaywall.org/}}, it contains an open database of 48,383,164 free scholarly articles that collect Open Access content from over 50,000 publishers and repositories. For the paper PDFs that are not too long, we use our own PDF parser to structure these PDF documents. Fourth, we filter the academic data from the Falcon-Refinedweb~\citep{refinedweb_penedo2023refinedweb} according to some domains. 
Generally, we believe the quality in the Falcon-Refinedweb~\footnote{\url{https://huggingface.co/datasets/tiiuae/falcon-refinedweb}} is good, what we need to do is to select out the high-quality academic data from it. 
Besides of the above-mentioned sources, we also use wiki pages (only English pages), bibliographic from Semantic Scholar~\footnote{\url{https://www.semanticscholar.org/}}, papers from PubMed.

In Table~\ref{tab:training_data}, we list the detailed information of the data we collected and used in this paper. Most of data are research papers, theses, and some other academic data.

\ 

\subsection{Modeling}
Neural network is essentially a \textit{function approximation} problem. Given a large amount of data $(\bx_i, \by_i) \ \text{for}\  i \in [1, N]$, our target is to learn a function $F(\cdot)$ to minimize the following loss function:
\begin{equation}
  loss =\frac{1}{N} \sum_{i=1}^{N} \mathcal{L}
 \left(F\left(\bx_{i} ;\bW\right), \by_{i}\right),
\label{eq:loss}
\end{equation}
where $\mathcal{L}(\cdot, \cdot)$ is the loss function.

After training stage, the inference is essentially an \textit{interpolation} process.

\textbf{Model Capability}. To expect the model will have strong approximation ability, we need to promise that the network will have a larger Lipschitz constant that is defined as:
\begin{equation*}
  \|F(\bx_1; \bW)-F(\bx_2; \bW)\| \leq L_0 \|\bx_1-\bx_2\|,
\label{eq:lipschitz}
\end{equation*}
where $L_0$ is the Lipschitz constant.

A large Lipschitz constant means that the model has a stronger nonlinearity, and thus it has a stronger approximation ability.
For instance, compared to the Convolution~\citep{convolution_lecun1998gradient} network, 
Transformer architecture~\citep{transformer_vaswani2017attention} has a much larger Lipschitz constant, and thus a powerful representation ability. Via understanding the Jacobian matrix of each module and its corresponding Lipschitz constant, we can theoretically estimate the representative ability of the network. Readers can refer to~\cite{understanding_dl_qi2023understanding} for a detailed analysis.

\textbf{Training Stability}. However, larger Lipschitz constant may lead to training instability. Thus, to promise the network will have a stable training process, we need to keep that

\begin{equation}
\bx^l, \frac{\partial \mathcal{L}}{\partial \bx^l} < \mathcal{R}, \ \text{for\ } l \in [1, L].
\label{eq:stability}
\end{equation}

The above equation means that the activations and their gradients should be bounded to the range of numerical representation (e.g., FP16, FP32 or BF16).

\textbf{Our Training Strategy}. 
To train AcademicGPT, we use the AdamW optimizer~\citep{adamw_loshchilov2017decoupled}, with $\beta_1$ and $\beta_2$ values set at 0.9 and 0.95 respectively and  $\epsilon = 10^{-8}$. We leverages a cosine learning rate schedule, decaying the final learning rate to be a mere $\frac{1}{10}$ of the peak learning rate 1.5e-5. Our used batch size is around 1.57M  tokens, where each sample comprises sequences of 4,096 tokens. For gradient accumulation, we accumulate 64 mini-batch. To train the model stably, we conduct the following tricks:
\begin{itemize}[leftmargin=*]
    \item we use BF16~\citep{bf16_kalamkar2019study} instead of FP16.
    \item we use FP32 for the LayerNorm~\citep{layernorm_ba2016layer} layer.
    \item we set gradient clipping to be 0.4 instead of 1.0.
    \item for the $\epsilon$ in the LayerNorm layer, we set it to be 1e-5.
    \item we use a longer warmup~\citep{warmup_loshchilov2016sgdr}.
\end{itemize}

These above tricks are either to extend the range $\mathcal{R}$ of numerical
representation or to constraint rapid growth of the Lipchistz constant $L_0$ of the modules or the whole network. In this way, we can promise Equation~\ref{eq:stability} will always hold true in the training process.

To speedup the training process, we also integrated some new and advanced techniques including FlashAttention2~\citep{flashattention2_dao2023flashattention} that not only speedup the attention module but also save a large amount of memory, Apex RMSNorm that implements a fused cuda kernel. Since AcademicGPT is a continual training model of LLaMA2-70B, it uses some same technology as LLaMA2 including RMSNorm~\citep{rmsnorm_zhang2019root} instead of LayerNorm, SwiGLU~\citep{swiglu_shazeer2020glu} instead of GeLU. For position embedding, it uses RoPE~\citep{rope_su2021roformer} instead of Alibi~\citep{alibi_alibi_press2021train}. For tokenizer, it uses BPE~\citep{bpe_sennrich2015neural}. It uses DeepSpeed~\citep{deepspeed_rasley2020deepspeed} with Zero~\citep{zero_rajbhandari2020zero}. Our training is based gpt-neox~\citep{neox_black2022gpt} framework in which we integrate many newly introduced skills. It takes around 37 days to finish the training of 120B data using 192 A100 GPUs with 40GB memory.

\ 

\subsection{Results}
We evaluate AcademicGPT on several benchmarks. First, we evaluate our models on some general benchmarks, including MMLU~\citep{mmlu_hendrycks2020measuring} and CEval~\citep{ceval_huang2023c}. Our goals are to evaluate whether the continual training will deteriorate the performance of the original LLaMA2~\citep{llama2_touvron2023llama} model and to evaluate the Chinese ability of our AcademicGPT after our continual training. Second, we evaluate the capability of AcademicGPT on some academic benchmarks, including PubMedQA~\citep{pubmedqa_jin2019pubmedqa}, SCIEval~\citep{scieval_sun2023scieval} and ComputerScienceQA. ComputerScienceQA is a newly created dataset by us to evaluate the capability of the model on computer science area. In default, when we mention LLaMA1 and LLaMA2, we mean LLaMA1-65B and LLaMA2-70B. 

\ 


\textbf{Results on MMLU}. We examine AcademicGPT's ability on MMLU. 
The MMLU test set covers 57 tasks including elementary mathematics, US history, computer
science, law, and more. To attain high accuracy on this test, models must possess
extensive world knowledge and problem solving ability. By comprehensively evaluating
the breadth and depth of a model’s academic and professional understanding,
MMLU can be used to analyze models across many tasks and to identify
important shortcomings.

Following some standard evaluation methods~\citep{llama_touvron2023llama, llama2_touvron2023llama, palm_chowdhery2022palm, openai_gpt4}, we use 5-shot setting for evaluation. In Table~\ref{tab:result_on_MMLU}, we report the averaged performance on 57 classes of MMLU test set, and compare AcademicGPT with LLaMA1 (65B), LLaMA2 (70B), ChatGPT (gpt-3.5-turbo-0613). In Table~\ref{tab:result_on_MMLU_subsets}, we show the results of AcademicGPT and LLaMA2 on several subjects in MMLU.

\begin{table}
\centering
\begin{tabular}{ c c  }
\hline
 Methods & Accuracy \\ \hline
 LLaMA-65B & 0.634 \\
 LLaMA2-70B & \textbf{0.693} \\  
 ChatGPT & 0.664 \\
 AcademicGPT & \textbf{0.688} \\
 \hline
\end{tabular}
\caption{Comparison of different methods on MMLU.}
\label{tab:result_on_MMLU}
\end{table}

We can find that the continual training on LLaMA2 will not deteriorate the performance averagely. Meanwhile, we observe that the results on several categories that have large amount of data used in our continual training will improve, but on some categories that is not largely covered in the data of our continual training, their performance will slightly decrease.

\begin{table}
\centering
\begin{tabular}{ c | c c | c c  }
\hline
 Methods & college\_computer\_science & college\_biology & high\_school\_geography &  sociology \\ \hline
 LLaMA2-70B & 0.58      &           0.813 & \textbf{0.889}   &     \textbf{0.881}\\
 AcademicGPT &  \textbf{0.62} &    \textbf{0.847} & 0.855 &      0.851\\
 \hline
\end{tabular}
\caption{Result comparison of AcademicGPT and LLaMA2 on some subjects in MMLU.}
\label{tab:result_on_MMLU_subsets}
\end{table}

\

\textbf{Results on CEval}.
To evaluate the capacity of AcademicGPT on Chinese language, we evaluate it on CEval benchmark, and compare it with several other methods.

\begin{table}[H]
\centering
\begin{tabular}{ c c  }
\hline
 Methods & Accuracy \\ \hline
 LLaMA-65B & 0.390 \\
 LLaMA2-70B & 0.508 \\  
 ChatGPT & 0.471 \\
 AcademicGPT & \textbf{0.551} \\
 \hline
\end{tabular}
\caption{Result comparison of different methods on CEval.}
\label{tab:result_on_ceval}
\end{table}

CEval is a Chinese evaluation toolkit, aiming to swiftly assess and understand a model's capabilities from various perspectives, especially its worldly knowledge and reasoning abilities. This assessment originates from real-world Chinese human exams spanning middle school, high school, university, and professional levels, covering 52 subjects including STEM, humanities, and social sciences. We utilize the valid set of CEval for evaluations during the model development process, which comprises 1,346 questions across all 52 subjects. 
During our assessment, we employ a 5-shot evaluation setting. The results are shwon in Table~\ref{tab:result_on_ceval}. 

We can see that from Table~\ref{tab:result_on_ceval}, by integrating a modest amount of Chinese common crawler content from textbooks and Baidu Baike (a Chinese version of Wikipedia), we enhanced the performance of AcademicGPT to 55.1\% on CEval from 50.8\% of the original LLaMA2. In our side-by-side evaluations, the Chinese-enhanced AcademicGPT significantly outperforms its original version in scenarios like academic reading assistance and translation.

\begin{table}[H]
\centering
\begin{tabular}{ c c  }
\hline
 Methods & Accuracy \\ \hline
 LLaMA-65B & 0.772 \\
 LLaMA2-70B & 0.776 \\  
 ChatGPT & 0.716 \\
 GPT-4-Base & 0.804 \\
 AcademicGPT & \textbf{0.806} \\
 \hline
\end{tabular}
\caption{Result comparison of different methods on PubMedQA. The result of GPT4 can be found at \url{https://pubmedqa.github.io/}. }
\label{tab:result_on_pubmedqa}
\end{table}

\

\textbf{Results on PubMedQA}.
PubMedQA~\citep{pubmedqa_jin2019pubmedqa}~\footnote{\url{https://pubmedqa.github.io/}} is a  biomedical question answering dataset that is collected from
PubMed abstracts. The task of PubMedQA is to answer research questions with three choices:
yes/no/maybe,  according to the corresponding abstracts. It consists of 1k expert-annotated, 61.2k unlabeled and 211.3k artificially generated QA
instances. Each instance is composed of four parts: a question, a context, a long answer and an assertion.
This question is either an existing research article title or content derived from the title. The context is
the corresponding abstract excluding its conclusion. The long answer is
the conclusion of the abstract and, presumably, answers the research question. Finally, there is a yes/no/maybe assertion which summarizes the conclusion.

For our evaluation, we only use the 1k expert-annotated instances. We use the 5-shot for our evaluation. The results of different models are shown in Table~\ref{tab:result_on_pubmedqa}.

We can see that from Table~\ref{tab:result_on_pubmedqa} on the PubMedQA dataset, our method achieved better results than LLaMA1, LLaMA2, ChatGPT3.5, and GPT4. We reckon this may be attributed to the presence of more medical-related corpus in our continual training data.

\

\textbf{Results on SCIEval}. SCIEval~\footnote{\url{https://github.com/OpenDFM/BAI-SciEval}}~\citep{scieval_sun2023scieval} is a scientific evaluation system based on Bloom's Taxonomy, designed to assess a model's performance in foundational knowledge, knowledge application, scientific computation, and research capabilities. The data primarily originates from Socratic Q\&A~\footnote{\url{https://socratic.org/socratic/contributing-to-socratic/asking-questions}} and integrates multiple public datasets, encompassing three subjects: biology, chemistry, and physics. We conducted tests using SCIEval's validation set, focusing solely on the objective questions within the validation set—a total of 1,187 questions, with 380 in biology, 643 in chemistry, and 164 in physics.

\begin{table}[h]
\centering
\begin{tabular}{ c c c c c }
\hline
Methods & Biology & Chemistry & Physics & Average Accuracy \\ \hline 
LLaMA1-65B  & 0.774 & 0.596 & 0.470 & 0.613\\ 
LLaMA2-70B  & 0.797 & 0.649 & 0.463 & 0.636 \\ 
ChatGPT & \textbf{0.813} & \textbf{0.705} & 0.518 & 0.679 \\ 
AcademicGPT & 0.800 & 0.680 & \textbf{0.585} & \textbf{0.688} \\ \hline
\end{tabular}
\caption{Result comparison of different methods on SCIEval.}
\label{tab:result_on_scieval}
\end{table}

We leverage 3-shot context learning evaluation as previous methods~\citep{scieval_sun2023scieval}. We compare AcademicGPT with ChatGPT3.5 and the original LLaMA2~\citep{llama2_touvron2023llama} that have been tested on SCIEval.
The results are reported in Table~\ref{tab:result_on_scieval}.

We can see that from Table~\ref{tab:result_on_scieval}, AcademicGPT improves the average accuracy from 63.6 obtained by LLaMA2 to 68.8, and also surpasses ChatGPT's score of 67.9. We can see that compared to ChatGPT, AcademicGPT performs better on physics but does not perform well on chemistry.

\

\textbf{Results on ComputerScienceQA}.
The ComputerScienceQA evaluation benchmark is built based on PapersWithCode~\footnote{\url{https://paperswithcode.com/}}. Papers With Code showcases trending Machine Learning research along with the corresponding implementation code. 

PapersWithCode consists of two sections: ``dataset'' and ``method''. The method part predominantly delves into descriptions of techniques detailed in research papers, whereas the dataset part pertains to dataset descriptions. This information is curated and reviewed by an open community. From a methodological viewpoint, PapersWithCode spans across seven major areas, each comprising multiple categorical layers. For instance, under ``Attention/Attention Mechanisms/Attention Patterns'', one would find descriptions of varied method concepts such as Strided Attention, Fixed Factorized Attention, Sliding Window Attention, etc. From a dataset perspective, PapersWithCode covers an array of modalities like Images, Text, Video, Audio, etc., providing a holistic and real-time overview of datasets in the Computer Science domain. Until to September 2023, our ComputerScienceQA includes 1,885 methods and 7,801 datasets.
Each sub-domain consists of several topics, and each topic contains an array of methodologies. For example, under ``self-attention'', there exists a multitude of distinct self-attention mechanism implementations, including linear attention, sparse attention, fast attention, dot-product attention, L2 similarity attention, etc., from a dataset angle.

Below, we will describe our construction strategy.
For the ``method'' question type:
\begin{enumerate}[leftmargin=*]
    \item Retrieve the method description and process it: case-insensitive matching of the description against the method's name and full\_name is done and replaced with ``()'', to prevent information leakage. All HTTP(s) links are removed to avoid data breaches.
    \item Craft the question prompt as:
Question: Which of the following options is a description of  ``{method.get(`full\_name', method[`name'])}''?
\end{enumerate}
The correct option stems from the method's description, while the distractor options are derived from the descriptions of other methods within the same domain collection.

\begin{figure}[H]
\begin{AIbox2}{{Four Samples in ComputerScienceQA \xspace}}
\parbox[t]{1.0\linewidth}
{
\tiny
\begin{alltt} 
{
{\large{\textbf{Case 1 (about method introduction):}}}

\textit{Question}: Which of the following options is a description of ``Convolution''? 

\textit{Choices}:\\
A: () softly switches the convolutional computation between different atrous rates and gathers the results using switch functions. The switch functions are spatially dependent, i.e., each location of the feature map might have different switches to control the outputs of (). To use () in a detector, we convert all the standard 3x3 convolutional layers in the bottom-up backbone to ().\\
B: A () is a type of matrix operation, consisting of a kernel, a small matrix of weights, that slides over input data performing element-wise multiplication with the part of the input it is on, then summing the results into an output.  Intuitively, a () allows for weight sharing - reducing the number of effective parameters - and image translation (allowing for the same feature to be detected in different parts of the input space).\\
C: A () layer is a simple extension to the standard convolutional layer. It has the same functional signature as a convolutional layer, but accomplishes the mapping by first concatenating extra channels to the incoming representation. These channels contain hard-coded coordinates, the most basic version of which is one channel for the $i$ coordinate and one for the $j$ coordinate. The () layer keeps the properties of few parameters and efficient computation from convolutions, but allows the network to learn to keep or to discard translation invariance as is needed for the task being learned. This is useful for coordinate transform based tasks where regular convolutions can fail.\\
D: While performs the channelwise and spatial-wise computation in one step, () splits the computation into two steps: applies a single convolutional filter per each input channel and is used to create a linear combination of the output of the depthwise convolution. The comparison of standard convolution and () is shown to the right.

\textit{Answer:}

{\large{\textbf{Case 2 (about method reference):}}}

\textit{Question}: which  of the following  paper proposed the method Bottleneck Residual Block? 

\textit{Choices}:\\
A: Deeper Learning with CoLU Activation\\
B: NEAT: Neural Attention Fields for End-to-End Autonomous Driving\\
C: Deep Residual Learning for Image Recognition\\
D: Few-shot Domain Adaptation by Causal Mechanism Transfer

\textit{Answer:}

{\large{\textbf{Case 3 (about dataset introduction):}}}

\textit{Question}: MNIST is a dataset which: 

\textit{Choices}:\\
A: Stanford Question Answering Dataset (SQuAD) into Spanish.\\ 
B: () is an open-source recreation of the [WebText](/dataset/webtext) corpus. The text is web content extracted from URLs shared on Reddit with at least three upvotes.\\
C: Maternal and Infant (It) Dataset is a large-scale dataset jointly labeled for classification, question answering and summarization in the domain of maternity and baby caring in Chinese. An entry in the dataset includes four fields: question (Q), description (D), class (C) and answer (A).\\
D: database) is a large collection of handwritten digits. It has a training set of 60,000 examples, and a test set of 10,000 examples. It is a subset of a larger NIST Special Database 3 (digits written by employees of the United States Census Bureau) and Special Database 1 (digits written by high school students) which contain monochrome images of handwritten digits".

\textit{Answer:}

{\large{\textbf{Case 4 (about dataset reference):}}}

\textit{Question}: which  of the following  paper introduced the dataset Multi-Genre Natural Language Inference (MultiNLI)? 

\textit{Choices}:\\
A: Hierarchical Text Generation and Planning for Strategic Dialogue\\
B: Hope Speech detection in under-resourced Kannada language\\
C: A Broad-Coverage Challenge Corpus for Sentence Understanding through Inference\\
D: Cards Against AI: Predicting Humor in a Fill-in-the-blank Party Game

\textit{Answer:}
}
\end{alltt}
}
\end{AIbox2}
\caption{Two QA samples from our ComputerScienceQA.}
\label{two_csqa_samples}
\end{figure}

Our methodology is primarily inspired by the the Galactica~\citep{galactica_taylor2022galactica} paper, aiming to gauge a model's proficiency in grasping methods and datasets within the computer science domain. The merits of such a construct method include 
\begin{itemize}[leftmargin=*]
    \item a comprehensive coverage of current mainstream knowledge in the CS domain and an objective, 
    \item a multiple-choice format that simplifies creation and facilitates accurate evaluation.
\end{itemize}

In conclusion, we have collated a total of 9,686 questions, of which 1,885 pertain to methods and 7,801 relate to datasets. Samples about method and dataset can be referred to in Figure~\ref{two_csqa_samples}.

\begin{table}[h]
\centering
\begin{tabular}{ c c c c c c}
\hline
 Methods & methods-intro & methods-refer & datasets-intro & datasets-refer & Overall \\ \hline
 LLaMA-65B & 0.519  & 0.882 & 0.608 & 0.829 & 0.710 \\
 LLaMA2-70B & 0.641 & 0.932 & 0.762 & 0.861 & 0.799 \\  
 ChatGPT & 0.715 & \textbf{0.937} & 0.753 & 0.839 & 0.811 \\
 AcademicGPT & \textbf{0.767} & 0.913 & \textbf{0.777} & \textbf{0.883} & \textbf{0.835} \\
 \hline
\end{tabular}
\caption{Result comparison of different methods on ComputerScienceQA.}
\label{tab:result_on_paperqa}
\end{table}

For our evaluation, we employed a three-shot approach. We contrasted our methodology with ChatGPT and the native architecture of LLaMA2. The outcomes can be viewed in Table~\ref{tab:result_on_paperqa}.

We can see that from Table~\ref{tab:result_on_paperqa} AcademicGPT performs much better than the original LLaMA2 and improves the performance from 79.9\% to 83.5\%. Compared to ChatGPT, it also shows better performance.

\

\section{Applications of AcademicGPT}
\label{sec:applications}
Based on AcademicGPT, we built several applications, including general academic question answering, AI-assisted paper reading, paper review, and AI-assisted title and abstract generation. In essence, by building upon the robust foundation of AcademicGPT, we not only enhance the capabilities of the model but also create several tools that can empower academic research. Figure~\ref{fig:overall_academicgpt_and_applications} shows our overall framework.

\

\subsection{General Academic Question Answering}
Academic question answering requires more rigors compared to general question answering. 
Our academic question answering is a LLM-empowered agent~\citep{mrkl_karpas2022mrkl, agent_weng2023prompt, agentsurvey_xi2023rise, toolformer_schick2023toolformer, visual_chatgpt_wu2023visual, huggginggpt_shen2023hugginggpt} which consists of the following modules: a AcademicGPT-powered engine that acts a brain, a planning and action module, memory, and tools. 
This system can harness the power of various academic tools, tailored to diverse types of questions such as paper retrieval, conceptual clarifications, or multi-paper comparisons and to paper recommendation.
 A overview of the AcademicGPT-empowered agent is shown in Figure~\ref{overview_of_chatscholar}. Below, we will  introduce each module in detail.

\begin{figure}[t]
\centering
\includegraphics[width=0.9\textwidth]{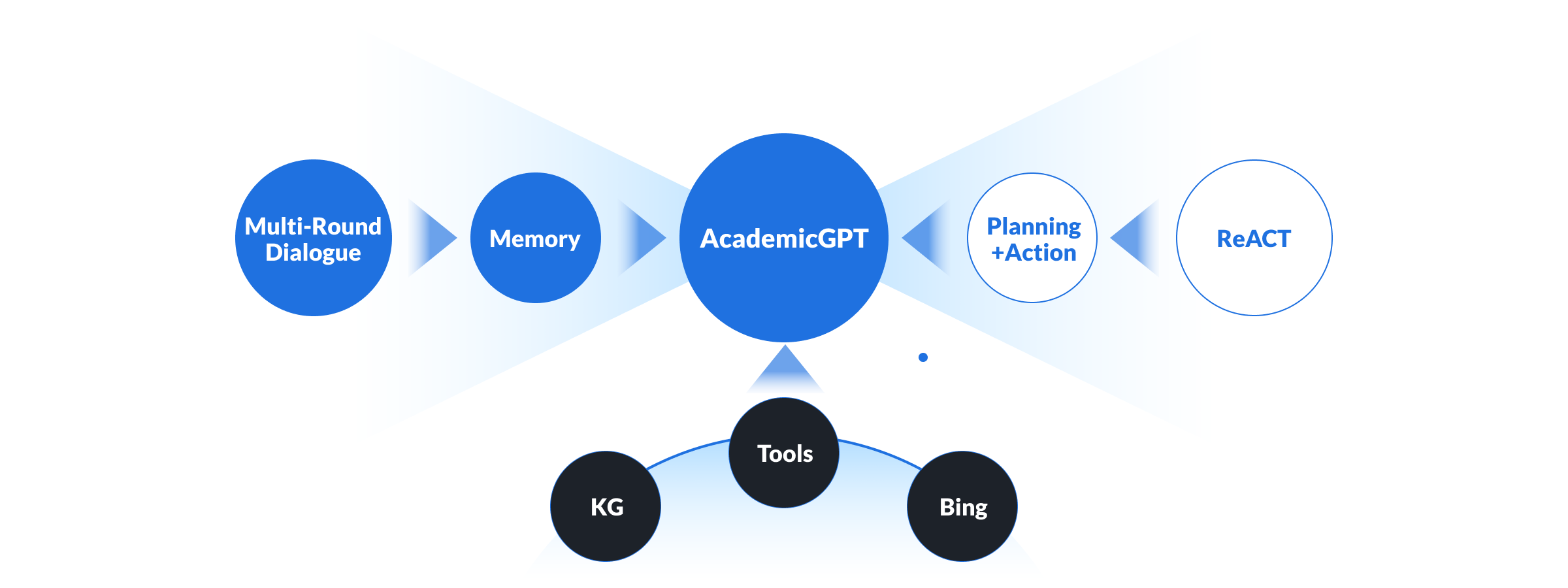}
\label{overview_of_chatscholar}
\caption{Overview of our AcademicGPT-powered academic question answering engine.}
\label{overview_of_chatscholar}
\end{figure}

\textbf{AcademicGPT-empowered engine}. As shown in Figure~\ref{overview_of_chatscholar}, the engine is the brain of the system. 
Essentially, our AcademicGPT-powered engine is a instruction-finetuned AcademicGPT. The engine should have the following two abilities:
\begin{itemize}[leftmargin=*]
    \item understanding and execute instructions
    \item knowing when to use tools, which tool to use, and how to use the tool.
\end{itemize}

To endow our model with the above-mentioned two abilities, our instruction-finetune data should include two types of data: general instruction-finetune data and instruction data for tool usage.  Our  instruction-finetune data primarily consists of our further cleaning of open-source data, including the cleaned Wizard~\footnote{\url{https://huggingface.co/WizardLM}}, LIMA~\footnote{\url{https://github.com/GaloisInc/LIMA}}, both Chinese and English versions of alpaca~\citep{alpaca}, and our constructed 384 tool usage instructions.

\textbf{Planning and Action}.
   Leveraging the capabilities of LLMs as the brain of our agent, the system can contemplate and strategize over diverse questions.  After LLMs, there are many works~\citep{cot_wei2022chain, selfconsistency_wang2022self, reflexion_shinn2023reflexion, react_yao2022react} focusing on improving models' planning and reasoning abilities including chain of thoughts (CoT)~\citep{cot_wei2022chain}, self-consistency~\citep{selfconsistency_wang2022self}, reflection~\citep{reflexion_shinn2023reflexion} and ReAct~\citep{react_yao2022react}.
   Our approach employed ReAct. ReAct expands the action space, combining discrete actions for specific tasks with linguistic constructs. This amalgamation seamlessly integrates reasoning and action into the LLM. 
   The ReAct method is a synthesis of reasoning and subsequent action. This method was conceptualized based on a keen observation of human behavior: humans tend to engage in a reasoning process between steps of multi-step tasks. We adapted this by enabling the LLMs to vocalize its ``inner monologue'', aligning subsequent actions with this articulated reasoning, thereby emulating human cognitive processes. This approach, tested across diverse datasets, achieved state-of-the-art results, boosting the credibility of LLMs and reducing its propensity for nonsensical outputs.

   Differing from ReAct~\citep{react_yao2022react}, our action outputs are in the JSON format, detailing the APIs used along with their respective parameters. Further insight into these parameters can be found in the following prompts as shown in Figure~\ref{fig:react_prompt}. 

\textbf{Memory}.
   All the historical multi-turn dialogues' contexts are considered as the model's short-term memory. In contrast, academic knowledge graphs retrieved via fuzzy keyword searches serve as long-term memory.

\textbf{Tool Utilization.} Many tools can be used in an agent including search engine, knowledge graph (KG), vector knowledge library, and other. In our system, we use the following tools: KG and Bing search engine.
For the KG, we use a elastic search (ES) based KG that incorporates information like author, title, abstract, publication date, institution, citations, and referenced papers into an ES setup, this tool offers fuzzy search capabilities across fields and logical sorting. Based on the KG, we add some features including recommendation of similar papers. This feature recommends multiple similar papers with precision, based on references and keywords.
For the Bing search engine, we also specially handled some websites, such as ``PapersWithCode''.
that allows for the retrieval of cutting-edge academic knowledge, such as the state-of-the-art results across datasets and their associated papers.
The detailed utility, application scenarios, and parameters of each API have been elaborated upon in the model's input prompts.

We have shown three cases in Figure~\ref{fig:general_academic_qa_case1}, Figure~\ref{fig:general_academic_qa_case2}, and Figure~\ref{fig:general_academic_qa_case3}. We can see that our system can do well on paper recommendation, concept explaining and etc.

\begin{figure}[H] 
\begin{AIbox1}{ReAct Prompt \xspace}
\vspace{-5px} 
\parbox{1\textwidth}
{
\scriptsize

\begin{alltt} 
{

\textbf{System Prompt:} \\
You are a pan-academic literature reading assistant. You can rigorously answer users' academic questions. You have access to the following tools:\\

\textbf{AcademicSearch}: \\
\{"description": "This is an tool for retrieving academic knowledge base through fuzzy matching on abstracts, authors, title, fieldOfStudy, publishDate or venue.", "input\_parameters": \{"abstracts": \{"type": "str", "description": "The query of the 
abstract. "\}, \\
"authors": \{"type": "list(str)", "description": "The authors of paper."\}, "fieldOfStudy": \{"type": "str", "description": "The field of the paper. "\}, \\ "publishDate": \{"type": "json", "description": "The key is gte or lte, and value is date(yyyy/MM/dd), such as \{\{'gte': '2020/01/01', 'lte': '2023/12/31'\}\}."\},  
"title": \{"type": "str", "description": "The title of paper. If there are multiple papers, use ';' to distinguish them, such as title1;title2."\}, \\
"venue": \{"type": "str", "description": "Published journals or conferences."\}, "sort\_by": \{"type": "json", "description": "The Key is abstracts, authors, fieldOfStudy, publishDate, title or venue. The value is 'desc' (descending) or 'asc' (ascending)."\}, "resultParameters": \{"type": "list(str)", "description": "Must required. Each item in the list should be abstracts, authors, fieldOfStudy, publishDate, title, venue or citationCount(the number of citations of the paper). \\
Format should be like ['xxx', 'xxx']"\}\}, "example of INPUT": "\{\{'title': 'xxx', 'resultParameters': ['authors', 'publishDate', 'abstracts']\}\}"\} \\

\textbf{WebSearchEngine}: \\
\{"description": "This is a web search engine. This tool will be very useful when you need to query basic academic knowledge and the latest academic knowledge.", "input\_parameters": \{"query": \{"type": "str", "description": "Must required. Input is the search query related to the question."\}\}, "example of INPUT": "\{\{'query': 'xxx'\}\}"\} \\

The way you use the tools is by specifying a Json blob.
Specifically, this Json should have a `action` key (with the name of the tool to use) and a `action\_input` key (with the input to the tool going here).

The only values that should be in the "action" field are: AcademicSearch, WebSearchEngine

The \$JSON\_BLOB should only contain a SINGLE action, do NOT return a list of multiple actions. 
\$JSON\_BLOB should start with '''. Here is an example of a valid \$JSON\_BLOB:

% '''
\{
\\  action: \$TOOL\_NAME,
\\  action\_input: \$INPUT \\
\}
% % % '''

ALWAYS use the following format:\\

Thought: you should always think about what to do \\
Action: \\
% % % '''
\$JSON\_BLOB
% % % ''' \\

Observation: the result of the action... (this Thought/Action/Observation can repeat N times) \\
Thought: I now know the final answer \\
Final Answer: the final answer to the original input question\\

\textbf{Input prompt:}\\
User: [User's utterence]\\
AI: [AI's utterence]

User: [User's utterence]



}
\end{alltt}
}
\vspace{-5px} 
\end{AIbox1} 
\caption{The ReAct prompt we used in our general academic question answering.}
\label{fig:react_prompt}
\end{figure}

\begin{figure}[H] 
\begin{AIbox}{{General Academic Question Answering (Case 1) \xspace}}
\vspace{-5px} 
\parbox{1\textwidth}
{\scriptsize
\begin{alltt} 
\textbf{Human:} 
What is the state-of-the-art performance on the CIFAR-10 dataset? Which paper does it come from?\\


\textbf{AI:} \hl{According to the provided information, the current state-of-the-art performance on the CIFAR-10 dataset is achieved by the paper "An Image is Worth 16x16 Words: Transformers for Image Recognition at Scale" with a percentage correct of 99.5 and 632M parameters. The paper was published on October 22, 2020.

Here is the full reference for the paper:

** An Image is Worth 16x16 Words: Transformers for Image Recognition at Scale ** 

Published: October 22, 2020

URL: https://paperswithcode.com/paper/an-image-is-worth-16x16-words-transformers-1

The paper proposes a novel approach to image recognition using transformers, achieving state-of-the-art performance on the CIFAR-10 dataset. The author introduces a new architecture that utilizes a combination of local and global context to improve the recognition of images. The paper also provides a thorough evaluation of the model's performance and compares it to other state-of-the-art methods.

It's worth noting that the performance of the model can be improved by using a larger dataset, pre-training the model on a larger dataset, or using a different architecture.}
\end{alltt}}
\vspace{-5px} 
\end{AIbox} 
\caption{The 1st case for General Academic Question Answering.}
\label{fig:general_academic_qa_case1}
\end{figure}

\begin{figure}[H] 
\begin{AIbox}{{General Academic Question Answering (Case 2) \xspace}}
\vspace{-5px} 
\parbox{1\textwidth}
{\scriptsize
\begin{alltt} 
\textbf{Human:} what are the currently best Large Language Models? \\

\textbf{AI:} \hl{
Currently, the best methods for large language models include:

\textbf{Meta AI's Llama}: Released in 2023, Llama is a large language model developed by Meta, with the largest version being 65 billion parameters in size.\\
\textbf{ChatGPT}: Developed by OpenAI, ChatGPT is a popular language model that has been used for a wide range of natural language processing tasks, including text generation, question answering, and language translation.\\
\textbf{Galactica}: Developed by Meta, Galactica is a large language model that has been used for a variety of tasks, including text generation and language translation.\\
\textbf{LaMDA}: Developed by Google, LaMDA is a large language model that has been used for a wide range of natural language processing tasks, including text generation, question answering, and language translation.\\
\textbf{GPT-3}: Developed by OpenAI, GPT-3 is a large language model that has been used for a wide range of natural language processing tasks, including text generation, question answering, and language translation.\\
\textbf{Megatron-Turing}:  Megatron-Turing is a large language model that has been used for a variety of tasks, including text generation and language translation.\\

It's worth noting that the field of large language models is rapidly evolving, and new models are being developed and improved upon regularly. These are some of the current state-of-the-art models, but there may be others that have been developed since the cutoff date of the knowledge base.}
\end{alltt}}
\vspace{-5px} 
\end{AIbox} 
\caption{The 2nd case for General Academic Question Answering.}
\label{fig:general_academic_qa_case2}
\end{figure}

\begin{figure}[H] 
\begin{AIbox}{{General Academic Question Answering (Case 3) \xspace}}
\vspace{-5px} 
\parbox{1\textwidth}
{\scriptsize
\begin{alltt} 
\textbf{Human:} can you please introduce the stable diffusion model in image generation? \\

\textbf{AI:} \hl{
Based on the latest research in image generation using stable diffusion, there have been several exciting developments in the field. Stable diffusion refers to a type of generative model that uses a diffusion process to model the generation of images. This approach has shown great promise in generating high-quality images that are both diverse and realistic.\\
\\
One of the most significant advancements in this field is the ability to generate images that are not only visually appealing but also semantically consistent. Researchers have demonstrated that by using stable diffusion, they can generate images that accurately represent the intended scene or object, while also capturing the nuances of the input data. This is particularly useful in applications such as computer vision, where the generated images need to be both visually realistic and semantically meaningful.\\
\\
Another area of progress is the use of stable diffusion for image-to-image translation tasks. By using stable diffusion, researchers have been able to translate images from one domain to another, such as converting a photo of a cat to a painting, while preserving the underlying structure and details of the original image. This has significant implications for applications such as image editing and enhancement.\\
\\
In addition, stable diffusion has also been used for image synthesis, where researchers have been able to generate completely new images that are not based on any existing data. This has been achieved by using stable diffusion to model the generation of images in a way that is both efficient and scalable. The resulting images are not only visually realistic but also diverse and coherent, opening up new possibilities for applications such as virtual reality and computer graphics.\\
\\
Furthermore, researchers have also explored the use of stable diffusion for image denoising and deblurring tasks. By using stable diffusion, they have been able to remove noise and blur from images, resulting in higher-quality images that are more suitable for analysis and processing.\\
\\
Overall, the latest results in image generation using stable diffusion show great promise for a wide range of applications, from computer vision to virtual reality. As the field continues to evolve, we can expect to see even more exciting developments and advancements in the coming years.
}
\end{alltt}}
\vspace{-5px} 
\end{AIbox} 
\caption{The 3rd case for General Academic Question Answering.}
\label{fig:general_academic_qa_case3}
\end{figure}

\

\subsection{AI-assisted Paper Reading}
AI-assisted paper reading is an effective tool to help readers and explain their confusions by an interactive QA format. 
For AI-assisted paper reading, LLMs with longer context windows are required because the full article of paper is usually long. However, training an LLM with long context windows from scratch requires significantly larger investments. In this technical report,
to achieve the ability of AI-assisted paper reading, we employ a two-stage pipeline. In the first stage, we continue the pre-training process and use a dynamic ntk-aware method to extend to larger window context. In the second stage, we use a supervised finetune mode and train the model on LongAlpaca-12k dataset. Below, we will describe these two stages in detail.

\begin{figure}[H] 
\begin{AIbox}{{AI-assisted Reading (Case 1) \xspace}}
\vspace{-5px} 
\parbox{1\textwidth}
{\scriptsize
\begin{alltt} 
\textbf{Context:} the content of this paper, or the content selected by the user from the paper.\\
Paper link: \url{https://arxiv.org/abs/1807.06521} \\

\textbf{Human:} Based on this paper, what are the future research directions?

\textbf{AI:} \hl{Based on the information presented in this paper, there are several areas for further research. \\
Firstly, the paper proposes a new attention mechanism for convolutional neural networks (CNNs) called Convolutional Block Attention Module (CBAM), which infers attention maps along two separate dimensions, channel and spatial, then the attention maps are multiplied to the input feature map for adaptive feature refinement. Future research could explore the effectiveness of this attention mechanism in other tasks and datasets, as well as the potential benefits of combining CBAM with other attention mechanisms or pre-training strategies. \\
Secondly, the paper validates the effectiveness of CBAM through extensive experiments on ImageNet-1K, MS COCO detection, and VOC 2007 detection datasets. Future research could explore the use of CBAM in other image classification and detection tasks, as well as the potential benefits of combining CBAM with other network architectures or optimization strategies. \\
Thirdly, the paper demonstrates that CBAM can be integrated into any CNN architectures seamlessly with negligible overheads and is end-to-end trainable along with base CNNs. Future research could explore the potential benefits of combining CBAM with other network architectures or optimization strategies, as well as the potential applications of CBAM in low-end devices or real-time systems. \\
Finally, the paper presents visualization results of the attention maps generated by CBAM, which indicate that CBAM can focus on target object regions better than other methods. Future research could explore the potential benefits of using attention visualization to interpret the behavior of neural networks and to improve their performance.
}
\end{alltt}}
\vspace{-5px} 
\end{AIbox} 
\caption{The 1st case for AI-assisted reading.}
\label{fig:ai_assisted_reading_case1}
\end{figure}

\begin{figure}[H] 
\begin{AIbox}{{AI-assisted Reading (Case 2) \xspace}}
\vspace{-5px} 
\parbox{1\textwidth}
{\scriptsize
\begin{alltt} 
\textbf{Context:} the content of this paper, or the content selected by the user from the paper.\\
Paper link: \url{https://arxiv.org/abs/2003.00418}
\\

\textbf{Human:} what is the workflow of LipGAN? %模型的工作流程
\\

\textbf{AI:} \hl{The workflow of the LipGAN model is as follows: \\

1. The input to the model is a face image, an audio signal, and a desired pose for the face.\\

2. The face image is encoded using a face encoder, and the audio signal is encoded using an audio encoder.\\

3. The encoded face image and audio signal are concatenated and fed into a face decoder to generate a talking face image that is in sync with the input audio signal. \\

4. The generated talking face image is fed into a discriminator network to determine whether it is synchronized with the audio signal.\\

5. The discriminator network is trained using a contrastive loss, while the generator is trained using an L1 reconstruction loss.\\

6. The model is trained in a GAN-like fashion, with the generator and discriminator competing against each other to improve the quality of the generated talking face images. \\

7. The final objective function of the model is a combination of the GAN objective function and the reconstruction loss. \\

8. During inference, the model generates the talking face image of the target speaker frame-by-frame using the input audio signal and the current frame as input.
}
\end{alltt}}
\vspace{-5px} 
\end{AIbox} 
\caption{The 2nd case for AI-assisted reading.}
\label{fig:ai_assisted_reading_case2}
\end{figure}

\begin{figure}[H] 
\begin{AIbox}{{Paper Review (Case 1) \xspace}}
\vspace{-5px} 
\parbox{1\textwidth}
{\scriptsize
\begin{alltt} 
\textbf{Title of the paper to be reviewed:} A comparison of dataset distillation and active learning in text classification \\ \textbf{URL}: \url{https://openreview.net/forum?id=UqmL1Oc4bCw}.


\textbf{Human:} You are a professional reviewer. You need to review a paper and discuss the novelty and originality of ideas, correctness, clarity, the significance of results, potential impact and quality of the presentation. \\
This is the paper for your review: <paper content> \\
% https://openreview.net/forum?id=UqmL1Oc4bCw

\textbf{AI:} \hl{Summary of the paper:\\
This paper studies the relationship between data distillation techniques and active learning approaches. Specifically, the authors conduct experiments on the Stanford Sentiment Tree corpus to demonstrate the superiority of data distillation compared to active learning.\\

Strengths:\\
* The paper is easy to follow.\\
* The topic is interesting.\\

Weaknesses: \\
* The writing needs to be polished. There are too many typos and grammar mistakes. \\
* The experimental setup is not convincing. First, there is no baseline provided. Second, the authors only conducted experiments on a single dataset. Third, the authors did not report the variance of the results. \\

Clarity, quality, novelty and reproducibility:See above.\\
Summary of the review:I recommend rejecting this paper due to the following reasons:\\
* The writing should be polished. Typos and grammar mistakes should be fixed.\\
* The experimental protocol is not convincing. No baseline is provided. Only one dataset is studied. Variance of the results is not reported.\\

Rating:strong reject
}
\end{alltt}}
\vspace{-5px} 
\end{AIbox} 
\caption{The 1st case for paper review.}
\label{fig:paper_review_case1}
\end{figure}

\begin{figure}[H] 
\begin{AIbox}{{Paper Review (Case 2) \xspace}}
\vspace{-5px} 
\parbox{1\textwidth}
{\scriptsize
\begin{alltt} 
\textbf{Title of the paper to be reviewed:} Coarse-to-fine Knowledge Graph Domain Adaptation based on Distantly-supervised Iterative Training \\ \textbf{URL}: \url{https://openreview.net/forum?id=-wDaB590pkt}.

\ 

\textbf{Human:} You are a professional reviewer. You need to review a paper and discuss the novelty and originality of ideas, correctness, clarity, the significance of results, potential impact and quality of the presentation. \\
This is the paper for your review: <paper content> \\
% https://openreview.net/forum?id=-wDaB590pkt

\textbf{AI:} \hl{Summary of the paper:\\
This paper proposes a Knowledge Graph Domain Adaptation Framework (KGDA) to construct a KG from fully unlabelled raw text data with the assistance of an existing KG in another domain. The key idea is to employ distant supervision to learn the mapping function from the input text to the KG space. The authors claim that their framework can discover new knowledge in the target domain.\\

**Strength**\\
1. The topic is interesting and important.\\

**Weaknesses**\\
1. My major concern is about the technical novelty. Most of the components in the proposed framework are borrowed from existing works. The most significant contribution seems to be the iterative training strategy. However, the authors do not provide sufficient evidence to support the superiority of the proposed method. \\
2. The writing needs to be improved. Some parts are hard to follow.\\
3. The experimental setup is not convincing. The authors did not compare with state-of-the-art baselines.\\

Summary of the review:My major concern is about the lack of technical novelty. Please refer to the Strength \& Weaknesses Section.\\

Rating:marginally below the acceptance threshold
}
\end{alltt}}
\vspace{-5px} 
\end{AIbox} 
\caption{The 2nd case for paper review.}
\label{fig:paper_review_case2}
\end{figure}

\begin{figure}[H] 
\begin{AIbox}{{Paper Review (Case 3) \xspace}}
\vspace{-5px} 
\parbox{1\textwidth}
{\scriptsize
\begin{alltt} 
\textbf{Title of the paper to be reviewed:} Assisted Learning for Organizations with Limited Imbalanced Data \\
\textbf{URL}: \url{https://openreview.net/forum?id=YqHW0o9wXae}.

\ 

\textbf{Human:} You are a professional reviewer. You need to review a paper and discuss the novelty and originality of ideas, correctness, clarity, the significance of results, potential impact and quality of the presentation. \\
This is the paper for your review: <paper content> \\
% https://openreview.net/forum?id=YqHW0o9wXae

\textbf{AI:} \hl{
Summary of the paper: \\
This paper proposes a new learning framework called "assisted learning" to address the issue of limited data availability in real-world applications. The key idea is to leverage the existence of a third-party service provider with abundant data to boost the learning performance of the learner. The authors provide theoretical analyses showing the convergence of the proposed approach. Empirical results on image classification and reinforcement learning demonstrate the effectiveness of the proposed approach.\\
Strengths:\\
1. The motivation of the paper is clear and easy to follow.\\
2. The proposed approach is simple and intuitive.\\
3. Theoretical analyses are provided to support the proposed approach.\\

Weaknesses:\\
1. My biggest concern about this paper is the lack of justification for the proposed approach. Why should we design a new learning framework instead of directly leveraging the pre-trained model from the service provider? What are the advantages of the proposed approach compared to simply fine-tuning the pre-trained weights from the service provider?\\
2. Another question is how to choose the hyperparameters $\gamma_L$ and $\gamma_P$. If $\gamma_L \ll \gamma_P$, then most of the data points come from the service provider. Doesn't it mean that the learner doesn't really benefit from the presence of the service provider? How to balance the trade-off between them?\\
3. The assumption that the learner and service provider have the same data distribution seems too strong. Is it possible to relax this assumption?\\
4. Since the focus of this paper is on helping the learner with limited data, why don't the authors compare the proposed approach with methods like knowledge distillation? Knowledge distillation is widely used to transfer knowledge from a teacher model to a student model with limited data.\\
5. The writing of the paper could be improved. For example, the authors should explain what $x_{t,i}$ means in Eq. (1).

}
\end{alltt}}
\vspace{-5px} 
\end{AIbox} 
\caption{The 3rd case for paper review.}
\label{fig:paper_review_case3}
\end{figure}

In stage 1, we use NTK~\citep{yarn_peng2023yarn, postional_interpolation_chen2023extending} to extent our window size to 32K, and we continue training on 5B sampled from our data collection as shown in Tabel~\ref{tab:training_data}. 
In stage 2, we use the LongAlpaca-12k~\citep{longalpac_long-alpaca, longlora} dataset for fully supervised finetune. The LongAlpaca-12k dataset comprises 9k long QA entries and an additional 3k short QA entries sampled from the original Alpaca dataset. This mix ensures that the model's proficiency in responding to shorter instructions remains unaffected. 
In line with the conventional Alpaca structure, the Long QA data adopts the following prompts for fine-tuning: 1) instruction: a string that lays out the task for the model. For instance, it might direct the model to answer a query after examining a segment of a book or a research paper. They have diversified the content and queries to ensure a wide range of instructions.
2) output: a string providing the response to the given instruction.

In engineering, we can also use some other methods to extend the window size. One choice is to train a small model, such as LLaMA-7B, to extract context information, and then use our AcademicGPT to generate the final answer.

In Figure~\ref{fig:ai_assisted_reading_case1} and Figure~\ref{fig:ai_assisted_reading_case2}, we have shown two cases to demonstrate the system of AI-assisted reading.

\

\subsection{Paper Review}
\textbf{Data Collection and Cleaning}.
The data for our paper review is from OpenReview~\footnote{\url{https://openreview.net/}}. We scraped 29,119 papers and 79,000 reviews from OpenReview. After that, we filtered out 7,115 papers that did not contain PDFs or review comments. 
Further, we removed some specific strings, such as ``Under review as a conference paper at ICLR 2023'' and ``Anonymous authors Paper under double-blind review'', and also deleted content from failed PDF parsing.
For the review cleaning, we removed reviews with excessive line breaks, those shorter than 100 tokens or longer than 2,000 tokens, and those inconsistent with the decision having the lowest confidence.

As Review Advisor~\citep{reviewadviser_yuan2022can}~\footnote{\url{https://github.com/neulab/ReviewAdvisor}},
we consider from seven aspects including ``clarity'',
``meaningful\_comparison'',
``motivation'',
``originality'',
``replicability'',
``soundness'',
``substance'', we use their open-source code to annotate the data.
Finally, we obtain 22,213 papers with 67,874 review comments for training and 500 papers with 1,513 review comments for testing. 

\begin{figure}[H]
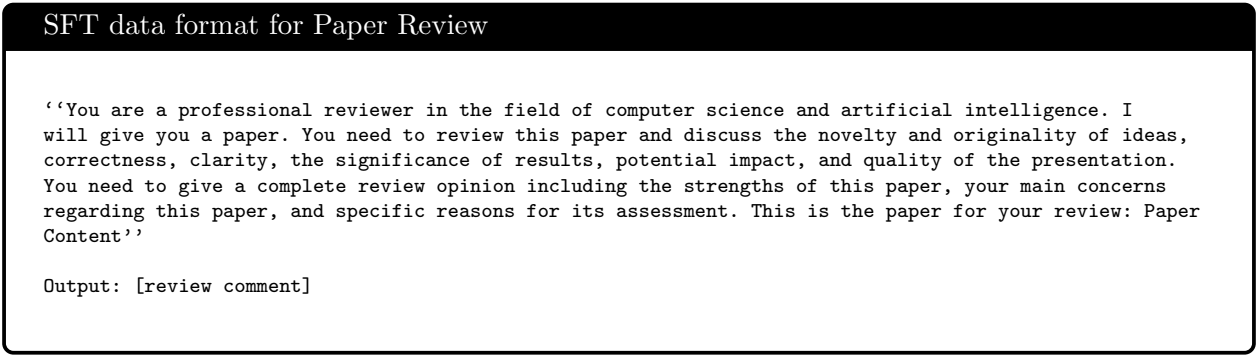
 
\begin{AIbox1}{{SFT data format for Paper Review \xspace}}
\vspace{-5px} 
\parbox{1\textwidth}
{
\scriptsize

\begin{alltt} 

``You are a professional reviewer in the field of computer science and artificial intelligence. I will give you a paper. You need to review this paper and discuss the novelty and originality of ideas, correctness, clarity, the significance of results, potential impact, and quality of the presentation. You need to give a complete review opinion including the strengths of this paper, your main concerns regarding this paper, and specific reasons for its assessment. This is the paper for your review: {Paper Content}''\\

Output: [review comment]

\end{alltt}
}
\end{AIbox1} 
\caption{SFT data format for Paper Review.}
\label{fig_sft_for_paperreivew}
\end{figure}

\textbf{SFT Details}. Our SFT data format is shown in Figure~\ref{fig_sft_for_paperreivew}.
We finetune the paper review model on AcademicGPT.
The maximum length is 8,192 instead of the original 4,096 in LLaMA2 and AcademicGPT. We use a batch size set of 128 and finetune 3 epochs.

\textbf{Metrics}. We employed three metrics for evaluation. The first is the accuracy of the final recommendation results; a recommendation is considered correct if it aligns with the meta-review recommendation, and incorrect otherwise. The second metric involves the accuracy related to the seven aspects mentioned earlier; if the aspects reported by us are also mentioned in the meta review, they are deemed accurate, otherwise incorrect. For instance, if out of $M$ predictions, $K$ are correct, the accuracy stands at $\frac{K}{M}$. The third metric is the recall rate. For example, if the meta mentions $N$ items for evaluation across all evaluation papers and we recall $L$, then the recall is $\frac{L}{N}$. The evaluation metrics used in this report are motivated by Review Advisor~\citep{reviewadviser_yuan2022can}.
The results are shown in Table~\ref{tab:result_of_paperreview}.

\begin{table}[h]
\centering
\begin{tabular}{ c c c c}
\hline
 Methods &  Final Recommendation Accuracy & Aspect Recall & Aspect Accuracy\\ \hline
 AcademicGPT & 68.4\% & 76.4\% & 24.8\%\\
 Human & 85.2\% & 81.6\% & 26.0\% \\
 \hline
\end{tabular}
\caption{Results of AcademicGPT and Human Reviewer on paper review. The accuracy and recall rates of Human Reviewers are determined based on their consistency with meta-reviewers.}
\label{tab:result_of_paperreview}
\end{table}

From  Table~\ref{tab:result_of_paperreview}, we can see that our final recommendation accuracy is 68\%. In the paper Review Adviser~\citep{reviewadviser_yuan2022can}, the authors concluded that the language models at the time did not possess the capability for paper reviewing. Our perspective is that the paper review script has certain reviewing capabilities and can serve as a reference comments for the authors. However, there is still a noticeable gap compared to human reviewers.

In Figure~\ref{fig:paper_review_case1}, Figure~\ref{fig:paper_review_case2}, and Figure~\ref{fig:paper_review_case3}, we show three cases of paper review.

\ 

\subsection{AI-assisted Title and Abstract Generation}

The essence of a research paper is often captured in its title and abstract. They not only provide a snapshot of the study's content but also play a pivotal role in garnering readers' attention. It is, therefore, paramount that they are crafted with precision and clarity.

\begin{figure}[H] 
\begin{AIbox}{{AI-assisted Title and Abstract Generation (Case 1) \xspace}}
\vspace{-5px} 
\parbox{1\textwidth}
{\scriptsize
\begin{alltt} 
\textbf{Input:} the introduction part from the paper ``DHOT-GM: Robust Graph Matching Using A Differentiable Hierarchical Optimal Transport Framework'', the paper url is \url{https://arxiv.org/pdf/2310.12081.pdf}.

\ 

\textbf{AI:} \\
\textbf{Title}:\hl{DHOT-GM: Differentiable Hierarchical Optimal Transport for Graph Matching}\\
\textbf{Abstract}:\hl{The task of graph matching attempts to find the node correspondence between two or more graphs, which is fundamental in graph-related applications.While most existing methods match graphs merely based on the adjacency matrices, or the node attributes, or the subgraph structures, they fail to consider the interactions between different modalities of graphs and are less robust in practice.To address these issues, we propose a multi-modal graph matching method, called DHOT-GM, based on a differentiable hierarchical optimal transport (DHOT) framework. Specifically, we capture the multi-modal information of graphs by a set of relational matrices. The matching results within different modalities and across modalities are measured by the lower-level optimal transport matrices. We further learn the significance of the modalities and the weights of the modality pairs by solving an upper-level optimal transport problem. The two-level optimal transport problems are solved iteratively, leading to the proposed DHOT framework. Compared with existing methods, our DHOT-GM method explicitly considers the matching results across different modalities and demonstrates their contributions to improving the final matching performance. Additionally, by learning the weights, our DHOT-GM method can find useful modalities and rely more on the reliable matching results. Extensive experiments on both synthetic and real-world graph matching tasks demonstrate the superiority of our method and its robustness in highly noisy cases.}

\ 

\textbf{Ground Truth}: \\
\textbf{Title}:DHOT-GM: Robust Graph Matching Using A Differentiable Hierarchical Optimal Transport Framework\\
\textbf{Abstract}:Graph matching is one of the most significant graph analytic tasks in practice, which aims
to find the node correspondence across different graphs. Most existing approaches rely on adjacency matrices or node embeddings when matching graphs, whose performances are often
sub-optimal because of not fully leveraging the multi-modal information hidden in graphs, such
as node attributes, subgraph structures, etc. In this study, we propose a novel and effective
graph matching method based on a differentiable hierarchical optimal transport (HOT) framework, called DHOT-GM. Essentially, our method represents each graph as a set of relational
matrices corresponding to the information of different modalities. Given two graphs, we enumerate all relational matrix pairs and obtain their matching results, and accordingly, infer the
node correspondence by the weighted averaging of the matching results. This method can be
implemented as computing the HOT distance between the two graphs — each matching result is
an optimal transport plan associated with the Gromov-Wasserstein (GW) distance between two
relational matrices, and the weights of all matching results are the elements of an upper-level
optimal transport plan defined on the matrix sets. We propose a bi-level optimization algorithm
to compute the HOT distance in a differentiable way, making the significance of the relational
matrices adjustable. Experiments on various graph matching tasks demonstrate the superiority
and robustness of our method compared to state-of-the-art approaches.

\end{alltt}}
\vspace{-5px} 
\end{AIbox} 
\caption{The 1st case for AI-assisted generation.}
\label{fig:ai_generation_title_case1}
\end{figure}

\begin{figure}[h] 
\begin{AIbox}{{AI-assisted Title and Abstract Generation (Case 2) \xspace}}
\vspace{-5px} 
\parbox{1\textwidth}
{\scriptsize
\begin{alltt} 
\textbf{Input:} the introduction part from the paper ``Fast and Low-Cost Approximate Multiplier for FPGAs using Dynamic Reconfiguration'', the paper url is \url{https://arxiv.org/pdf/2310.10053.pdf}.

\ 

\textbf{AI:} \\
\textbf{Title}:\hl{DyRecMul: Low-Power Reconfigurable Approximate Multiplier for Machine Learning Inference on FPGAs}\\
\textbf{Abstract}:\hl{Approximation techniques can accelerate multiplication operations by improving efficiency in terms of delay, energy consumption, and circuit area, which is critical for applications such as machine learning that require a large number of multiplications.This letter proposes DyRecMul, a dynamically reconfigurable approximate multiplier, which is optimized for machine learning inference on AMD-Xilinx FPGAs.It utilizes approximate 3-2 and 4-2 compressors in a reconfigurable partial product accumulation tree.DyRecMul is optimized for low-cost multiplication of short bit-width operands.It also addresses the limited dynamic range of such operands by transforming fixed-point numbers to a floating-point representation.The letter presents the design details of an INT8 version of the multiplier and its evaluation results.We demonstrate that DyRecMul offers a 64\% reduction in delay and 49\% reduction in energy consumption, while the accuracy loss is negligible in inference tasks.}


\ 

\textbf{Ground Truth}: \\
\textbf{Title}:Fast and Low-Cost Approximate Multiplier for FPGAs using Dynamic Reconfiguration\\
\textbf{Abstract}:Multipliers are widely-used arithmetic operators in
digital signal processing and machine learning circuits. Due to
their relatively high complexity, they can have high latency and
be a significant source of power consumption. One strategy to
alleviate these limitations is to use approximate computing. This
paper thus introduces an original FPGA-based approximate multiplier specifically optimized for machine learning computations. It
utilizes dynamically reconfigurable lookup table (LUT) primitives
in AMD-Xilinx technology to realize the core part of the computations. The paper provides an in-depth analysis of the hardware
architecture, implementation outcomes, and accuracy evaluations
of the multiplier proposed in INT8 precision. Implementation
results on an AMD-Xilinx Kintex Ultrascale+ FPGA demonstrate
remarkable savings of 64\% and 67\% in LUT utilization for signed
multiplication and multiply-and-accumulation configurations, respectively, when compared to the standard Xilinx multiplier core.
Accuracy measurements on four popular deep learning (DL)
benchmarks indicate a minimal average accuracy decrease of less
than 0.29\% during post-training deployment, with the maximum
reduction staying less than 0.33\%. The source code of this work
is available on GitHub.

\end{alltt}}
\vspace{-5px} 
\end{AIbox} 
\caption{The 2nd case for AI-assisted generation.}
\label{fig:ai_generation_title_case2}
\end{figure}

In our technical report, inspired by the readers' understanding process, we designed our data formulation as: 
``Introduction \textit{Experiments}  \textit{Results}  
\textless begin\_generate\textgreater Title:content;Abstract:content.''. Where the content in \textit{Italy} font is optional. ``\textless begin\_generate\textgreater'' is a special token to trigger the generation process.

To ensure robustness in our approach, we drew data from an extensive corpus of 1 million research papers. Our methodology employed a supervised training mode, harnessing the power of labeled data to guide our model's learning. 

We present two generation cases in Figure~\ref{fig:ai_generation_title_case1} and Figure~\ref{fig:ai_generation_title_case2}. These figures showcase the model's ability to generate coherent and relevant titles and abstracts based on new test data, underscoring the potential of our approach in aiding the academic community.

\section{Conclusion}
\label{sec:conclusion}
In this technical report, we have illuminated two principal advancements we made in the realm of academic research.
Firstly, we introduce AcademicGPT, A LLM tailored specifically for academic research. Trained on a colossal 120 billion tokens, it underscores the potential of extensive academic datasets, ensuring a high degree of precision in grasping scientific subtleties.
Secondly, we have taken AcademicGPT's capabilities further by applying it in a range of applications, from a nuanced General Academic Q\&A system to AI-assisted reading and content creation. Our Q\&A tool, empowered by the ReAct framework, enriches academic dialogues by maintaining context. Furthermore, our initiatives in simplifying dense academic texts and in reviewing papers position AI as an indispensable tool for researchers. Notably, the adaptability our AI showcases in content generation, like abstracts, highlights its versatility. 
In conclusion, AcademicGPT and its associated applications represent a pioneering leap in bridging advanced AI technologies with the demands of academic research. Through these endeavors, we anticipate a substantial shift in how information is processed, interacted with, and generated within the academic sphere.

\bibliographystyle{apalike}
\bibliography{main}

\newpage
\appendix
\addtocontents{toc}{\protect\setcounter{tocdepth}{2}}

\end{document}